\documentclass[10pt, a4paper]{article}

\usepackage{lrec-coling2024} 

\usepackage{times}
\usepackage{latexsym}
\usepackage[english]{babel}
\usepackage[T1]{fontenc}
\usepackage[utf8]{inputenc}
\usepackage{microtype}
\usepackage{blindtext}
\usepackage{array}
\usepackage{cleveref}
\usepackage{booktabs}
\usepackage{multirow}
\usepackage{graphicx}
\usepackage{graphics}
\usepackage{subfig}
\usepackage{makecell}
\usepackage{xpatch}

\title{Towards Algorithmic Fidelity: Mental Health Representation across Demographics in Synthetic vs. Human-generated Data}
\name{Shinka Mori, Oana Ignat, Andrew Lee, Rada Mihalcea} 

\address{University of Michigan, Ann Arbor\\ Ann Arbor, MI, USA\\\{shinkamo, oignat, ajyl, mihalcea\}@umich.edu\\}

\abstract{
Synthetic data generation has the potential to impact applications and domains with scarce data. However, before such data is used for sensitive tasks such as mental health, we need an understanding of how different demographics are represented in it.
In our paper, we analyze the potential of producing synthetic data using GPT-3 by exploring the various stressors it attributes to different race and gender combinations, to provide insight for future researchers looking into using LLMs for data generation. Using GPT-3, we develop  \textsc{HeadRoom}, a synthetic dataset of 3,120 posts
about depression-triggering stressors, by controlling for race, gender, and time frame (before and after COVID-19). 
Using this dataset, we conduct semantic and lexical analyses to (1) identify the predominant stressors for each demographic group; and (2) compare our synthetic data to a human-generated dataset. 
We present the procedures to generate queries to develop depression data using GPT-3, and conduct analyzes to uncover the types of stressors it assigns to demographic groups, which could be used to test the limitations of LLMs for synthetic data generation for depression data.
Our findings show that synthetic data mimics some of the human-generated data distribution for the predominant depression stressors across diverse demographics. 
 \\ \newline \Keywords{Synthetic Data Generation, LLMs, Bias in LLMs, Mental Health Datasets} }

\begin{document}

\maketitleabstract

\section{Introduction}
The emergence of Large Language Models (LLMs) poses many exciting use cases in various applications~\citep{Bang2023-bz}. 
In particular, synthetic data generation \citep{Tang2023-eu} has great potential to impact domains such as mental health, where it can scale hard-to-acquire data to improve medical information extraction~\citep{agrawal_large_2022}, provide clinical decision support~\citep{shen_chatgpt_2023}, and enhance patient-doctor communication~\citep{kreimeyer_natural_2017}.

\renewcommand{\arraystretch}{2}
\begin{table}[h] 
\small
\centering
\resizebox{\columnwidth}{!}{%
\begin{tabular}{p{3cm}|p{4cm}} 
  Prompt Template & Sample Output\\ 
 \midrule
 ``I want you to act like a \{race\} \{gender\} who is feeling depressed. Write a \textit{blog post} to describe the main source of stress in your life' & \textit{.. I'm \textbf{not good enough}, like I'm not doing enough. I'm struggling to \textbf{make ends meet} and I'm constantly worried about \textbf{money}. I'm worried about \textbf{my family} and their \textbf{safety} ..}.\\ 
 \midrule

\end{tabular}
}
\caption{
Example of prompt templates used for \textsc{HeadRoom}, as well as sample outputs.
}
\label{tab:prompts}
\end{table}
\renewcommand{\arraystretch}{1}
\vspace{-0.05cm}

However, before using synthetic data, we need to understand the potential biases across demographics within the underlying models generating such data. 
Otherwise, a subsequent model trained on biased synthetic data can have undesirable consequences, such as misrepresentations of minority voices or, specifically in mental health, a misdiagnosis~\cite{Potts_Burnam_Wells_1991, Call_Shafer_2018}.

To address this need, we conduct extensive analyses to understand the similarities and differences between synthetic and human-generated data.
We focus on mental health data, specifically depression stressors across races and genders.
We study if GPT-3 accurately captures depression stressors across demographics and if the stressors map closely to those found in human-generated data.
\citet{argyle_out_2022} coin the term ``algorithmic fidelity'' to describe the degree to which models mimic the real-life distributions for a particular group.
Inspired by this, we aim to measure how accurately GPT-3 represents depression stressors for different demographics with the following research questions:
\begin{description}
\item [RQ1:]
What are the depression stressors identified by GPT-3 for different demographic groups and does it capture demographic biases?
\item[RQ2:] How does synthetic data about depression stressors compare to human-generated data across demographics?
\end{description}

We closely follow the analyses done by \citet{aguirre_using_2022} on human-generated data to discover patterns of depression stressors among demographics.
Namely, we generate a similar dataset by prompting GPT-3 to produce outputs representative of diverse demographics and compare our findings to theirs.

Our work makes the following contributions.
\textbf{First}, we develop and publish \textsc{HeadRoom}: a synt\textsc{HE}tic d\textsc{A}taset of Depression-triggering st\textsc{R}ess\textsc{O}rs acr\textsc{O}ss de\textsc{M}ographics, using GPT-3 while controlling for race, gender, context, and time phase -- before and after COVID-19.
\textbf{Second}, we identify the most predominant depression stressors for each demographic group.
\textbf{Third}, we conduct semantic and syntactic analyses to compare our synthetic data to a human-generated dataset. 
Our findings show that GPT-3 exhibits some degree of ``algorithmic fidelity'' -- the generated data mimics some real-life data distributions for the most prevalent depression stressors among diverse demographics.

\section{Related Work}

\paragraph{LLMs for Generating Mental Health Datasets Across Demographics.}
Psychological studies show that depression affects racial and gender groups differently~\cite{Brody_Pratt_Hughes_2018}.
Despite this, there are still discrepancies where minority groups are often overlooked for depression diagnoses~\cite{Stockdale_Lagomasino_Siddique_McGuire_Miranda_2008}. 
While demographic information is a key aspect to consider when conducting mental health studies, obtaining such data in the mental health domain is challenging due to safety and privacy regulations~\cite{mattern-etal-2022-differentially}. 
As a result, researchers often turn to alternative methods of obtaining demographic labels, such as using automated classifiers, keywords, or lists of names~\cite{wang_its_2018}.
However, as presented in ~\citet{field_survey_2021}, such methods fail to account for the multidimensionality of race due to simplifications inherent in classification models:
i.e., classifiers predicting demographics in tweets perform poorly on Asian and Hispanic samples~\citep{wood-doughty_using_2021}.
Furthermore, commonly used mental health datasets, such as CLPsych~\citep{coppersmith_clpsych_2015} and Multitask~\citep{benton_multi-task_2017}, underrepresent specific demographics such as men and Hispanic individuals~\citep{aguirre_gender_2021}. 

An alternative to predicting demographic labels using machine learning is to \emph{generate} demographic data using LLMs.
\citet{argyle_out_2022} show that GPT-3 can generate political stances regarding recent elections in the United States that strongly correlate with real-life voter distributions. 
\citet{moller_is_2023} compare the performance of classifiers trained on human-generated versus LLM-generated data, demonstrating that classifiers trained on synthetic data can perform well on tasks such as social dimensions.

Considering the inherent risks of applying these approaches to mental health tasks, measuring the \textit{algorithmic fidelity} of LLMs in the mental health domain is essential.
For instance, \citet{lin_gendered_2022} demonstrate that LLMs carry different mental health stigmas for men and women.
In our work, we also study demographic biases in LLMs by generating synthetic data using GPT-3 and analyzing it against human-generated data.

\paragraph{Depression-triggering Stressors Across Demographics Analysis.}

Depression stressors can vary greatly depending on the demographic, due to systemic racism, racial dynamics, gender discrimination, immigration status, and other factors such as COVID-19~\citep{mcknight-eily_racial_2021}.
Specifically, \citet{loveys_cross-cultural_2018} analyze data from self-reported depression users in an online peer support community.\footnote{\url{https://www.7cups.com/}}
Similar to our work, \citet{loveys_cross-cultural_2018} use Linguistic Inquiry and Word Count (LIWC)~\cite{Pennebaker2007LinguisticIA, Pennebaker2015TheDA}, a lexical analysis tool, to compare the stressors between racial groups, and found some critical differences in stressor patterns between demographics.  

\section{Datasets}


To answer our research questions, we generate data using GPT-3 while controlling for race, gender, and time (before and after COVID-19), to simulate human-generated data.
We then conduct semantic and lexical analyses to find patterns in the synthetic data.
Next, we compare our findings to those based on The University of Maryland - OurDataHelps dataset (UMD-ODH) ~\cite{kelly_blinded_2020, kelly_can_2021}, a demographically diverse human-generated dataset about depression stressors.

\subsection{UMD-ODH: Human-generated Data}
UMD-ODH~\citep{kelly_blinded_2020, kelly_can_2021} contains open-ended responses from patients clinically diagnosed with depression and psychosis.
Patients were asked: \textit{``Describe the biggest source of stress in your life at the moment. What things have you done to deal with it?''}

\citet{aguirre_using_2022} further process this data by selecting the survey responses that have demographic data available, 
resulting in $2,607$ survey responses.
The resulting demographic information is shown in \Cref{tab:demographics}.
\begin{table*}[h]

    \centering
    \scalebox{0.85}{
    \begin{tabular}{l l l l l l l l l l l}
    \toprule
    \multirow{1}{*}{} & \multicolumn{5}{c}{\textsc{Race}} & \multicolumn{3}{c}{\textsc{Gender}} & \multicolumn{2}{c}{\textsc{COVID-19}} \\
    \cmidrule(lr){2-6} \cmidrule(lr){7-9} \cmidrule(lr){9-11}
      Responses & White & Black & Asian  & Latinx & Other & Male & Female & Other & Before Pandemic & After Pandemic\\
    \midrule
    UMD-ODH & 1,761 & 221 & 246 & 277 & 102 & 1,857 & 659 & 94 & 890 & 1,717\\
    \textsc{HeadRoom} & 780 & 780 & 780 & 780 & -- & 1,500 & 1,500 & -- & 1,440 & 1,680 \\
    
    \midrule
   
    \end{tabular}
  
    }
   \caption{Demographic statistics from UMD-ODH and \textsc{HeadRoom}.
}
\label{tab:demographics}
\end{table*}

\begin{table*}[h] \small
    \begin{tabular}{c|p{5cm}| p{5cm} | c}
    Topic & UMD-ODH & \textsc{HeadRoom} & Similarity \\
        \toprule
         Family & family, focus, year, planning, friends
         &
         can, stress, deal, lot, person
         & 0.78
         \\
        \midrule 
         Work &  work, lot, hard, week, balance
         & 
         
         job, stress, work, sourc, life, work
         & 0.90
         \\
         \midrule 
         Health & 
         like, feel, surgery, found, productive
         &
         constant, feel, take, health, mental
         & 0.85 \\
        \midrule
         Finance & 
        money, bills, pay, sleep, lack
        & job, struggl, find, make, end
        & 0.94
        \\         \midrule

         Relationship & 
         day, stressed, relationship, tried, think
         &
         feel, thing, make, depress, like
         & 0.92\\        \midrule
         School & 
         problems, friends, program, plans, dissertation
         &
         asian, succeed, expect, pressur, fall
         
         & 0.90\\        \midrule
         News, Social media & 
         help, people, social, use, avoid
         &
         like, climat, current, stress, polit
         & 0.84\\        \midrule
         Unemployment & 
         job, new, finding, lost, looking
         & look, lost, time, get, month
         & 0.87\\
         
         %
         %

    \end{tabular}
    \caption{The keywords corresponding to each overarching topic in UMD-ODH and \textsc{HeadRoom}, together with the cosine similarity between the averaged GloVe embeddings~\cite{pennington-etal-2014-glove} of the keywords corresponding to each topic.  
    A cosine similarity between two sets of random words gives us a baseline of $0.75$.
    }
    \label{tab:keywords}
\end{table*}
\subsection{\textsc{HeadRoom}: GPT-3 generated Data}
We generate our synthetic dataset with GPT-3.\footnote{\texttt{Text-Davinci-003}, \url{https://platform.openai.com/docs/models/gpt-3-5}} 
We use GPT-3 because it is one of the largest LLMs available, and has been demonstrated to effectively emulate human texts \citep{argyle_out_2022}, but our study can be done with any LLM.\footnote{We also attempted to use ChatGPT, but due to its content filters, the prompt had to be heavily engineered, which may add confounding variables.}

\paragraph{Prompt Tuning.}
To simulate human-generated data, we paraphrase the prompt that \citet{kelly_blinded_2020, kelly_can_2021} use for their \textit{human survey}. 

We find that we can obtain more detailed responses with additional context in our prompts.
Therefore, we provide additional context such as \textit{writing a blog post}, \textit{posting on Reddit},\footnote{r\textbackslash Depression} or \textit{talking to a therapist}.
We use three contexts to obtain more diverse responses.
For each prompt, we also specify the user's \textit{gender} (women and men), \textit{race} (Asian, African American, Hispanic, and White), and \textit{context} (blog post, Reddit post, and therapy session).
We produce outputs for each \textit{race}, \textit{gender}, and \textit{context} combination. 
The prompts and example outputs are displayed in \Cref{tab:prompts}. 

The human-generated data also contains samples collected after the start of COVID-19, which may affect the stressor patterns. Therefore we also control for time by indicating the year (2020, 2021) in the prompts while preserving the data distribution.  Prompts that do not indicate the year are assumed to represent pre-COVID-19 samples.

For the \textit{blog post} context, we generate $720$ samples, $30$ samples per demographic group before COVID-19 and $60$ samples after COVID-19.
For the other two contexts, we generate $2,400$ total samples, $150$ samples per demographic group.
The data statistics are summarized in \Cref{tab:demographics}.


\begin{table}
    \centering
    \scalebox{0.87}{
    \begin{tabular}{l l p{1cm} l p{1cm} }
    \multicolumn{5}{ c }{\sc Gender} \\
    \toprule
       &Category & Ratio & Category & Ratio\\ 
       \midrule
       {(a)} & \multicolumn{2}{c}{Women (+)} & \multicolumn{2}{c}{Men (-)}\\
    \midrule
        &female & 3.95 & male & -4.06 \\
        &i  & 2.83 & we & -1.88 \\ 
	&pro1 & 2.60 & verb & -1.76 \\ 
	&ppron  &1.32 & tentat  & -1.21 \\ 
	&home & 1.10 & money & -1.02\\

    \midrule
    \multicolumn{5}{ c }{\sc Ethnicity} \\
      \midrule
        &Category & Ratio & Category & Ratio\\ 
       \midrule
     {(b)}&\multicolumn{2}{c}{Asian (+)} & \multicolumn{2}{c}{African American (-)}\\
    \midrule
	&work & 4.06 & see & -8.41 \\ 
	&nonflu & 3.11 & percept & -4.71  \\ 
	&home & 2.31 & health &  -3.88\\ 
        &reward & 1.57 & and & -1.76 \\ 
	&achiev & 1.53 &  compare & -1.68\\
    \midrule
     {(c)} & \multicolumn{2}{c}{Asian (+)} & \multicolumn{2}{c}{White (-)}\\
    \midrule
	&leisure & 2.73 & anx & -2.60 \\ 
	&work & 2.24 & focusfuture & -1.50 \\ 
	&reward &  2.14 & tentat & -1.37 \\ 
	&achiev & 1.62 & ingest & -1.32 \\
        &negate& 1.34 & relativ & -0.99\\
    \midrule
     {(d)} & \multicolumn{2}{c}{Hispanic (+)} & \multicolumn{2}{c}{White (-)}\\
    \midrule
	&home & 7.48 & insight & -3.36 \\ 
	&leisure &  7.37& percept & -3.17 \\ 
	&family&  7.18 & cogproc & -2.95 \\ 
	&affiliation & 5.30 &  see  & -2.62 \\ 
	&social & 2.51 & compare & -1.83   \\
\midrule
     {(e)} & \multicolumn{2}{c}{African American (+)} & \multicolumn{2}{c}{White (-)}\\
    \midrule
	&see & 5.88 & insight & -2.28 \\ 
	&bio & 4.01 & adverb & -2.25  \\ 
	&percept & 3.78 &  tentat & -2.06 \\ 
	&health &  3.40 & you & -1.80 \\ 
	&body & 2.78 & space & -1.78 \\ 
 \midrule
     {(f)} & \multicolumn{2}{c}{Hispanic (+)} & \multicolumn{2}{c}{African American (-)}\\
   \midrule
	&home & 7.33 & see & -8.49 \\
	&family & 6.79 & percept & -6.96 \\
	&leisure & 6.72 & bio & -5.24 \\ 
	&affiliation & 3.67 & health & -4.67 \\
	&social & 3.51 & feel & -4.40 \\
\midrule
     {(g)} & \multicolumn{2}{c}{Hispanic (+)} & \multicolumn{2}{c}{Asian (-)}\\
\midrule
	&affiliation & 5.13 & cogproc & -3.00 \\ 
	&home & 5.04 & i & -2.75 \\ 
	&leisure & 4.70 & reward &  -2.60 \\ 
	&family & 4.61 & negate & -2.39 \\ 
	&social &  2.73 & certain & -2.23 \\ 

\end{tabular}
}
\caption{Highlights of LIWC log-odds ratio analysis on \textsc{HeadRoom} showing LIWC categories related to predominant stressors when comparing between genders and demographic group pairs.  For the (+) group, higher score indicates higher prominence; for the (-) group, lower score indicates higher prominence
}
\label{tab:liwc-small}
\end{table}

\section{Dataset Analysis Methods}

Following \citet{aguirre_using_2022}, we conduct two analyses on our synthetic dataset: (1) Semantic analyses using Structural Topic Model (STM), and (2) Lexical analyses using log-odds-ratio with Latent Dirichlet prior. 

\paragraph{Semantic Analyses.}
STM  is a variant of Latent Dirichlet allocation (LDA) that also allows the addition of covariates, or metadata, to accompany the textual features.
Unlike LDA, which calculates topic prevalence and content from Dirichlet distributions whose parameters are set in advance, STM uses metadata to find the topic prevalence and content. 
Following \citet{aguirre_using_2022} who annotated and filtered their topics to $25$,  we use gender, race, and time (before and after COVID-19) and generate $25$ topics.
Two annotators labeled the topics based on their most prevalent keywords, while filtering out unclear topics.
The annotators obtained a Fleiss’ kappa score of $k=0.52$, which shows a \textit{moderate} agreement~\cite{Fleiss1971MeasuringNS, Fleiss1973StatisticalMF}.

We obtain $23$ fine-grained topics that we manually cluster into eleven overarching topics.
The fine-grained and corresponding overarching topics can be seen in the Appendix (\Cref{tab:topics}).
Of the eleven overarching topics, eight match those in the human-generated data.
The matching topics and their keywords are shown in \Cref{tab:keywords}. 
Topics from UMD-ODH that did not have a match with our data include \texttt{school/ grad school} and \texttt{daily stress}.  UMD-ODH has two topics related to \texttt{school} -- the first relates to school in general, and the second relates to graduate school.
While our dataset has a topic for \texttt{general stress}, we concluded that none of the keywords are similar enough to be considered a match.\footnote{\texttt{feeling stuck, staying strong, uncertainty, comparing to others, helplessness, stress and anxiety, loneliness, perfectionism}}

The four overarching topics that appear in the synthetic data and not in the human-generated data are: \texttt{general stress, racism and police violence, immigration status} and \texttt{pandemic}.
We conduct a pairwise analysis between each gender and race pair using these topics.  Effectively, to find the difference between topic proportions for each demographic pair, we estimate a regression to find the topic proportion with the added covariate information.  This is then used to extract the prevalence of a topic (topic distribution) for each demographic pair.  We show and discuss the difference in the topic prevalence for each demographic pair in \Cref{fig:topic_diff} and \Cref{fig:topic_diff_appendix}. 


\paragraph{Lexical Analyses.}
LIWC~\cite{Pennebaker2007LinguisticIA, Pennebaker2015TheDA} includes dictionaries of English words related to human cognitive processes. 
Specifically, we use the LIWC 2015 dictionary, which contains $6,400$ word stems. Each word stem is assigned to multiple categories, e.g., \textit{father} is assigned to: \texttt{male, family} and \texttt{social}.

\citet{aguirre_using_2022} apply log-odds-ratio with Latent Dirichlet prior, based on the work of \citet{monroe_fightin_2017}, which aims to capture how a demographic group uses a specific LIWC category compared to another demographic group.
For example, to compare the proportions in which one group uses the \texttt{negemo} (negative emotion) LIWC category compared to another group, we calculate the log-odds-ratio to get the odds of \texttt{negemo} being used in the first group compared to the latter.
To calculate the Dirichlet prior, we use the LIWC category counts in the CLPsych dataset~\citep{coppersmith_clpsych_2015}.
Note that we do not normalize the results with a control text unrelated to depression, to preserve comparison fidelity with \citet{aguirre_using_2022}, who also do not normalize.


We show the top five words that have a high log-odds-ratio in \Cref{tab:liwc-small} and highlight the LIWC categories also present in the pairwise lexical analysis from \citet{aguirre_using_2022} in the Appendix Tables \ref{tab:liwc_wm}, \ref{tab:liwc_aw}, \ref{tab:liwc_hw}, \ref{tab:liwc_bw}, \ref{tab:liwc_hb}, \ref{tab:liwc_ha}, and \ref{tab:liwc_ab}. 
Insights from the data analysis are presented in \Cref{sec:results}.

\section{Research Questions} \label{sec:results}

\paragraph{RQ1. What are the depression stressors identified by GPT-3 for different demographic groups, and do they capture demographic biases?}

\begin{figure*}[]
\centering
\begin{tabular}{c c}
  \includegraphics[width=70mm]{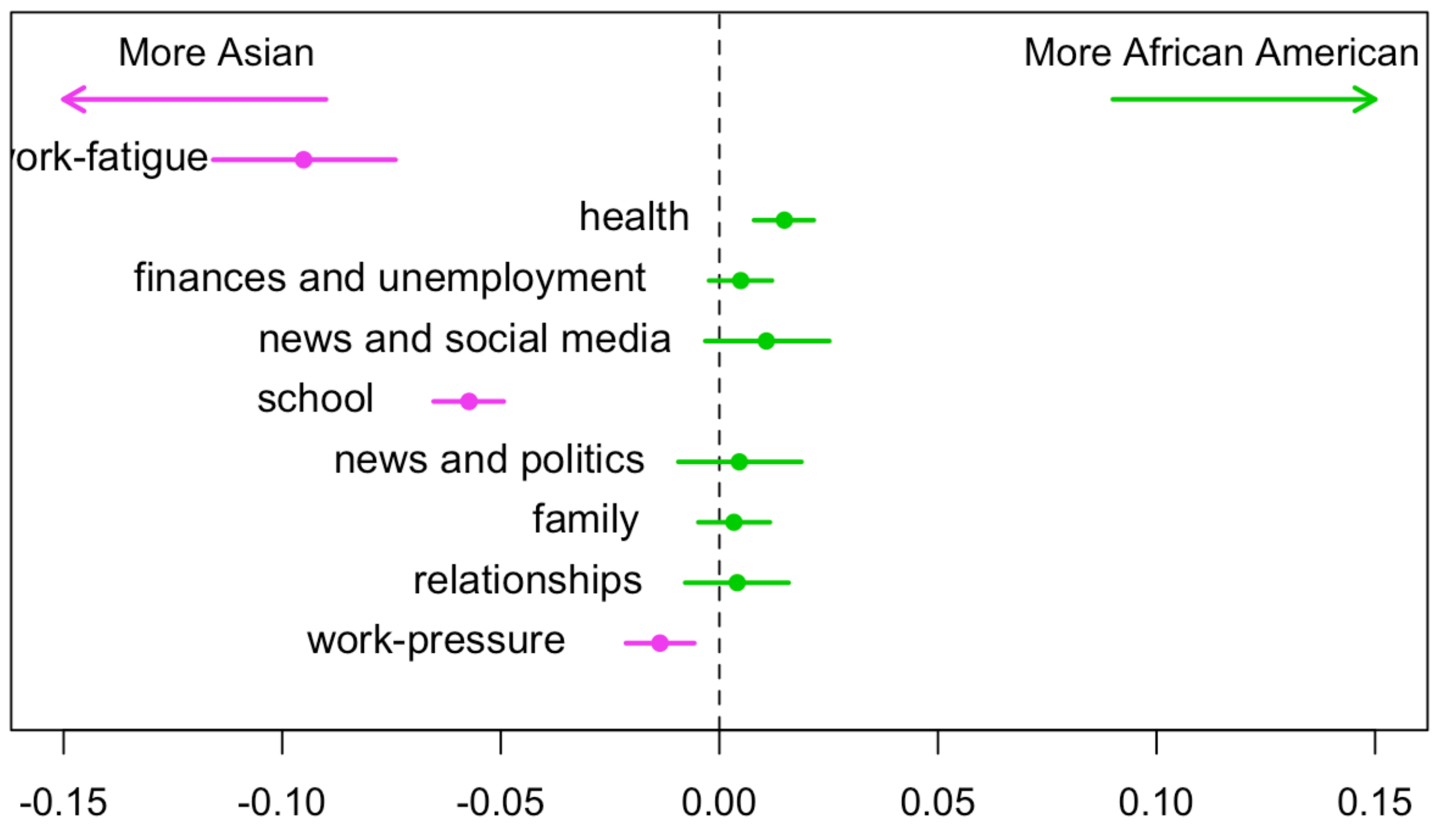} &   \includegraphics[width=70mm]{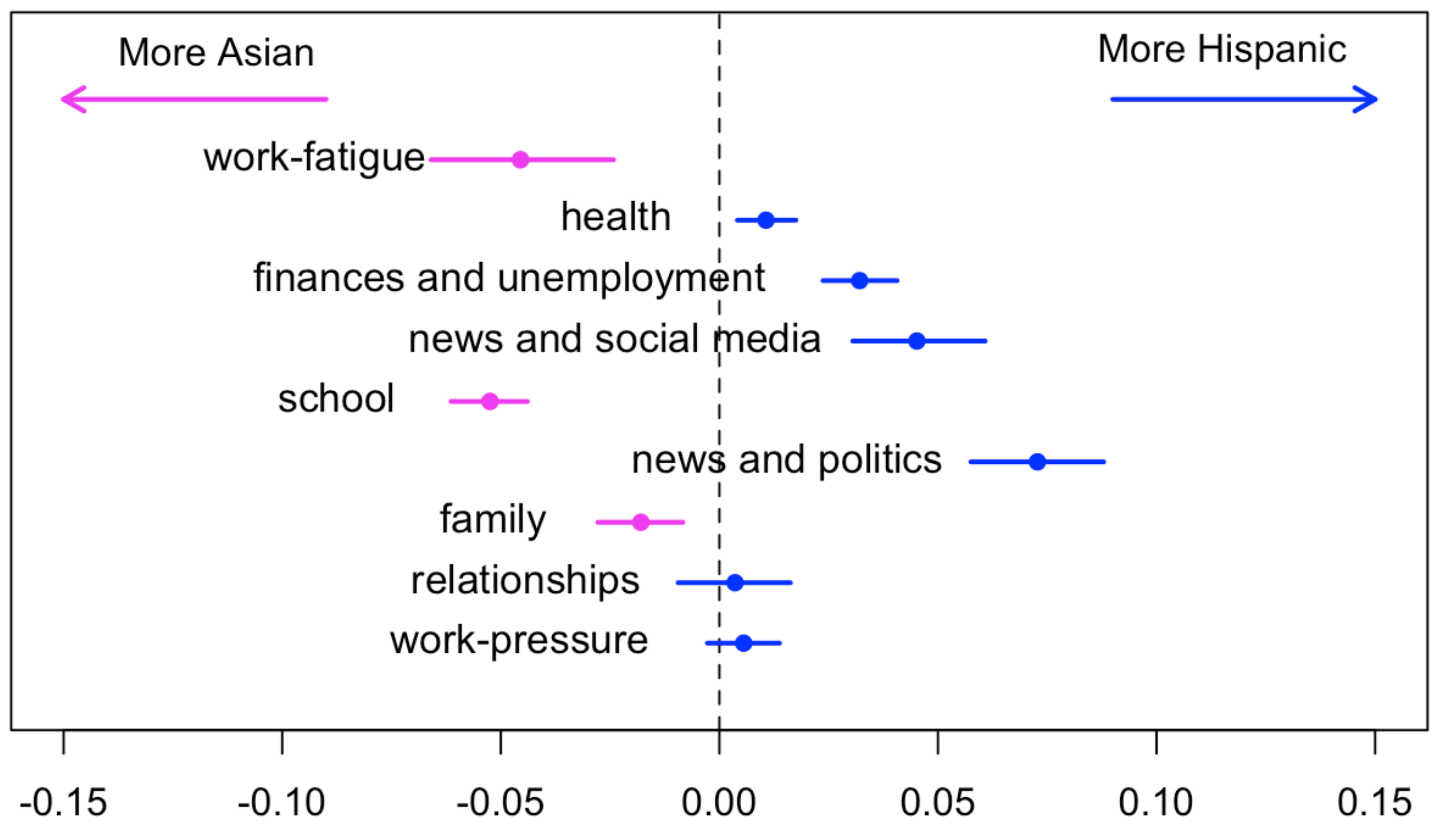} \\
(a) Asian vs. African American & (b) Asian vs. Hispanic \\[10pt]
 \includegraphics[width=70mm]{figures/a_b.pdf} &   \includegraphics[width=70mm]{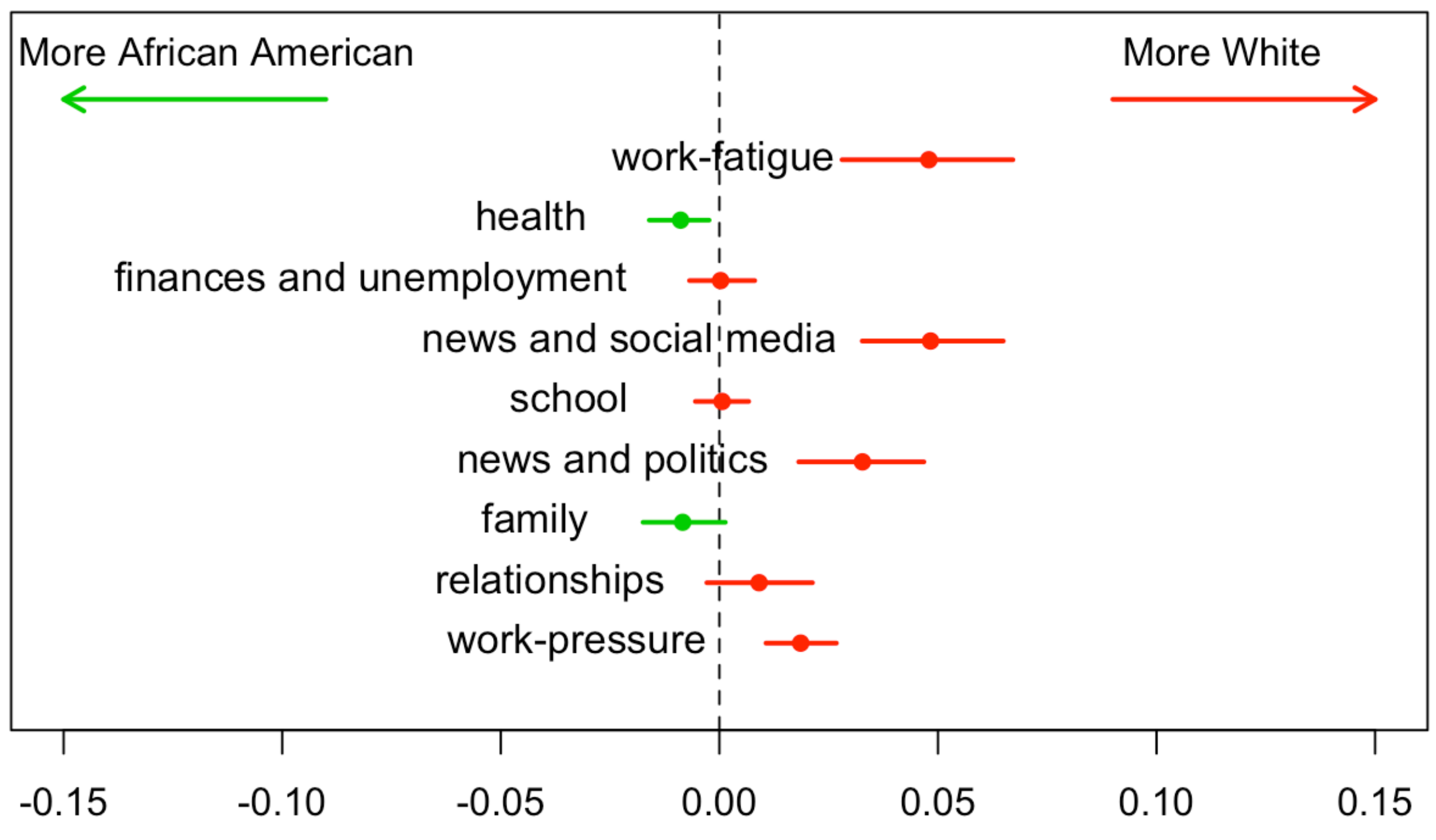} \\
(c) Asian vs. White & (d) African American vs. White \\[10pt]
\includegraphics[width=70mm]
{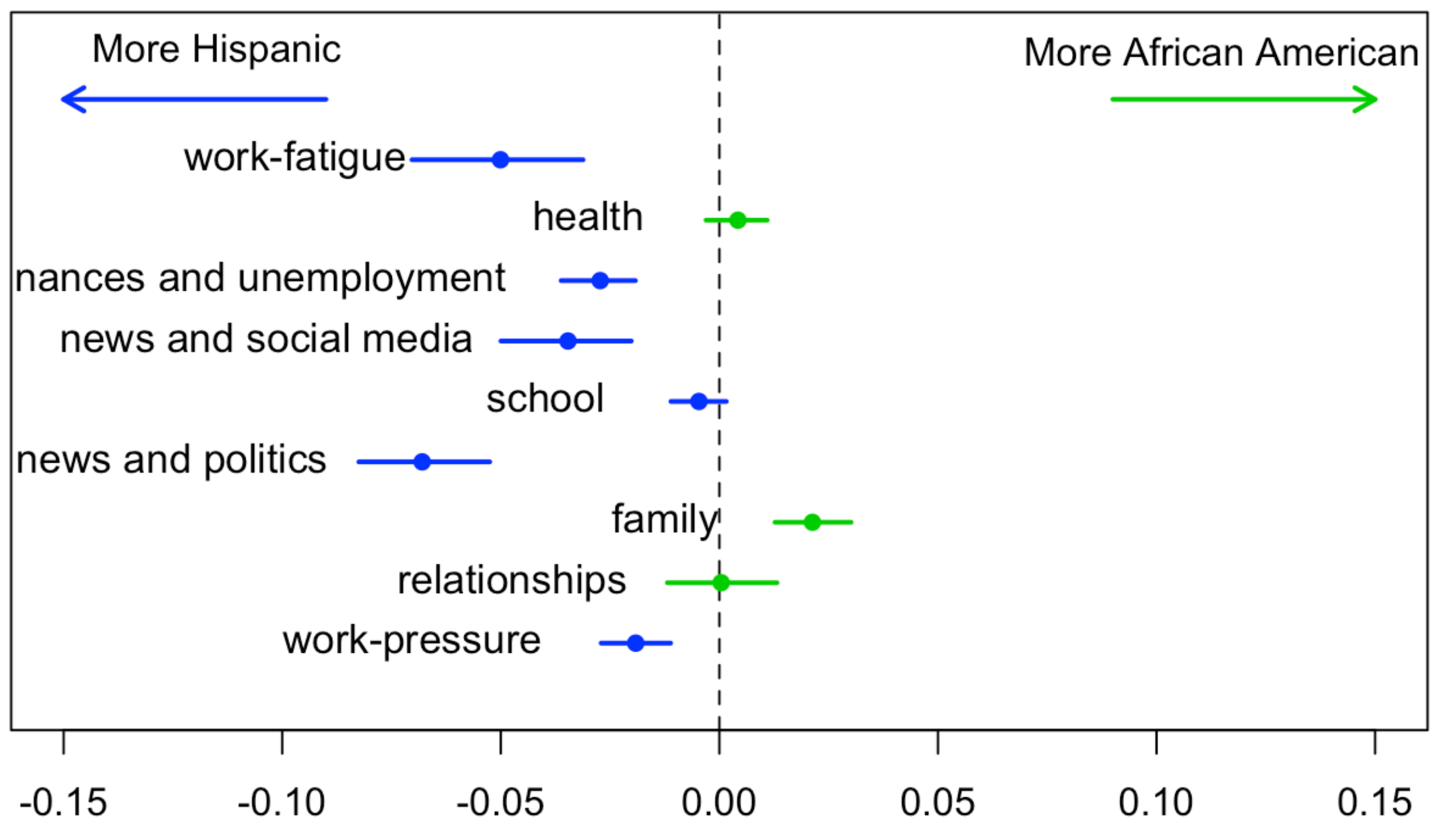} &   \includegraphics[width=70mm]{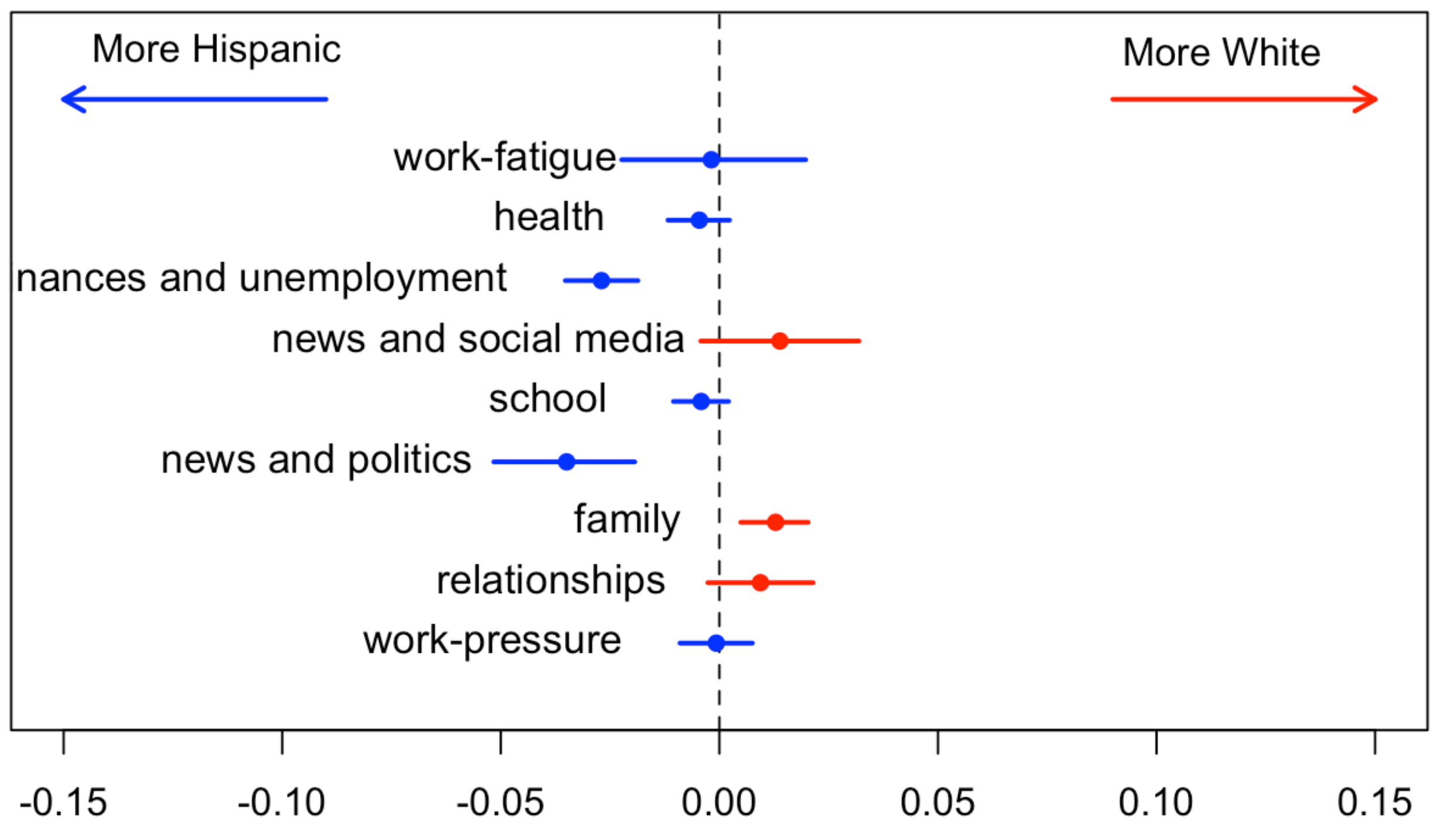} \\
(e) Hispanic vs. African American & (f) Hispanic vs. White \\[10pt]
\multicolumn{2}{c}{\includegraphics[width=70mm]{figures/b_comp.pdf} \includegraphics[width=70mm]{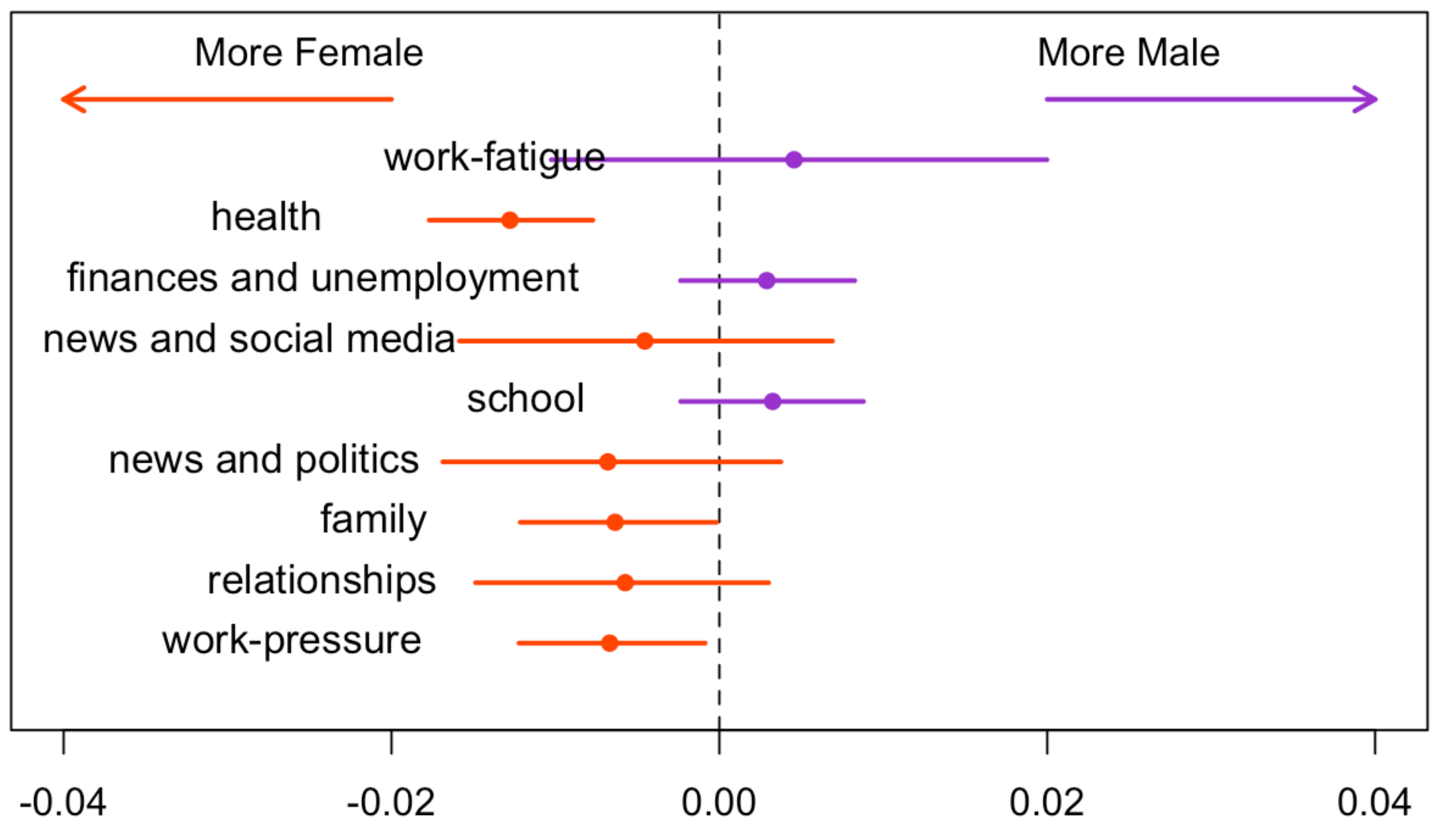} }\\
\multicolumn{2}{c}{(g) Women vs. Men}
\end{tabular}
\caption{
Topic Modeling: topic proportion between different demographics, as detected in GPT-generated data and in real-life data. 
Colors represent different races and genders: Men -- purple, Women -- orange, Asian -- magenta, African American -- green, Hispanic -- blue, and White -- red.
The bars represent confidence intervals. The closer to the graph extremities, the more prevalent the topics for the corresponding demographics. For example, graph (a) Asian vs. African American shows that stressors such as \texttt{work1/ work-fatigue}, \texttt{work2/ work-pressure} and \texttt{school} are more prevalent for Asian than for African American. \textit{Best viewed in color.}}
\label{fig:topic_diff}
\end{figure*}
\begin{figure*}[]
\centering
\begin{tabular}{c c}
  \includegraphics[width=70mm]{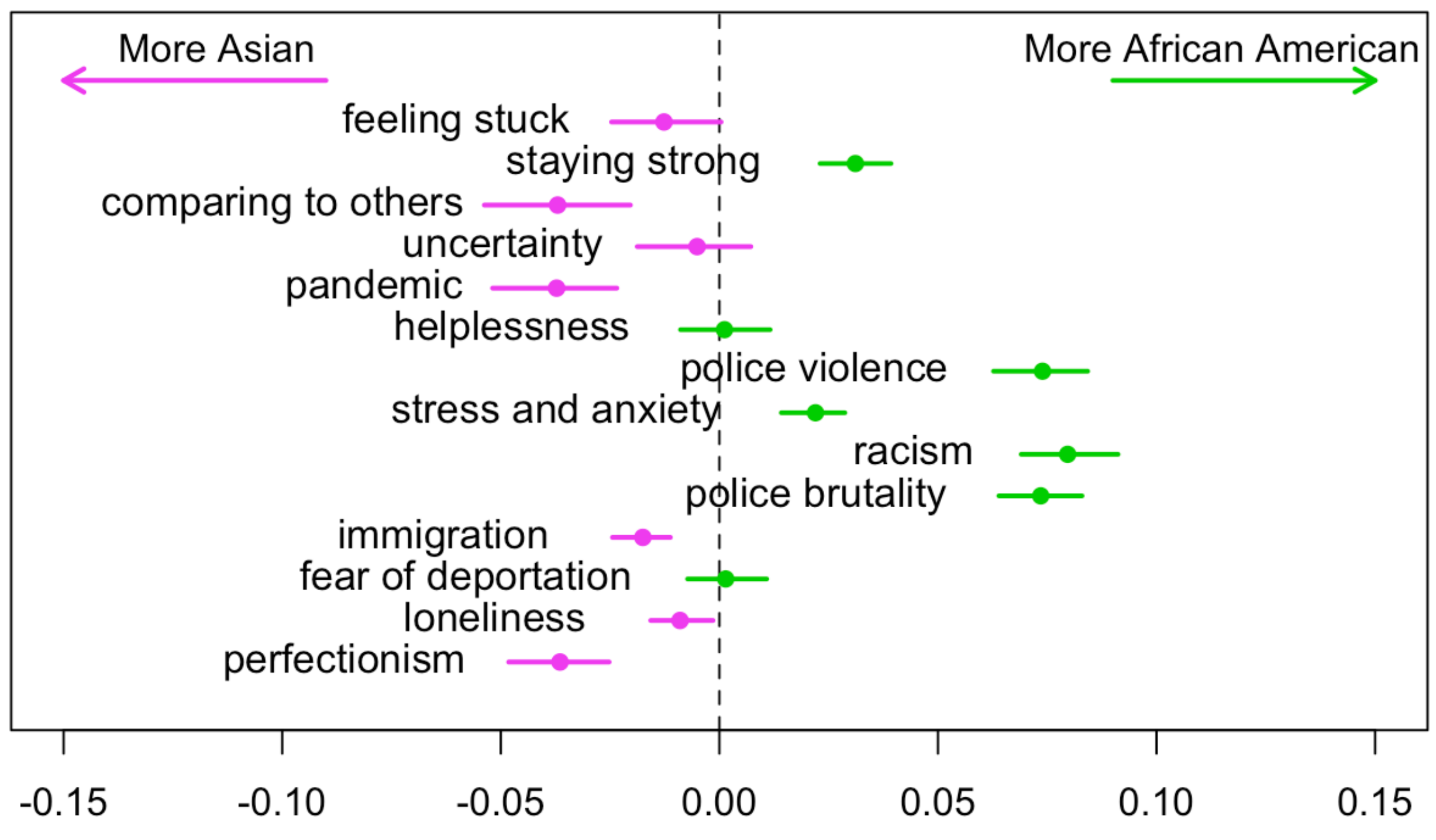} &   \includegraphics[width=70mm]{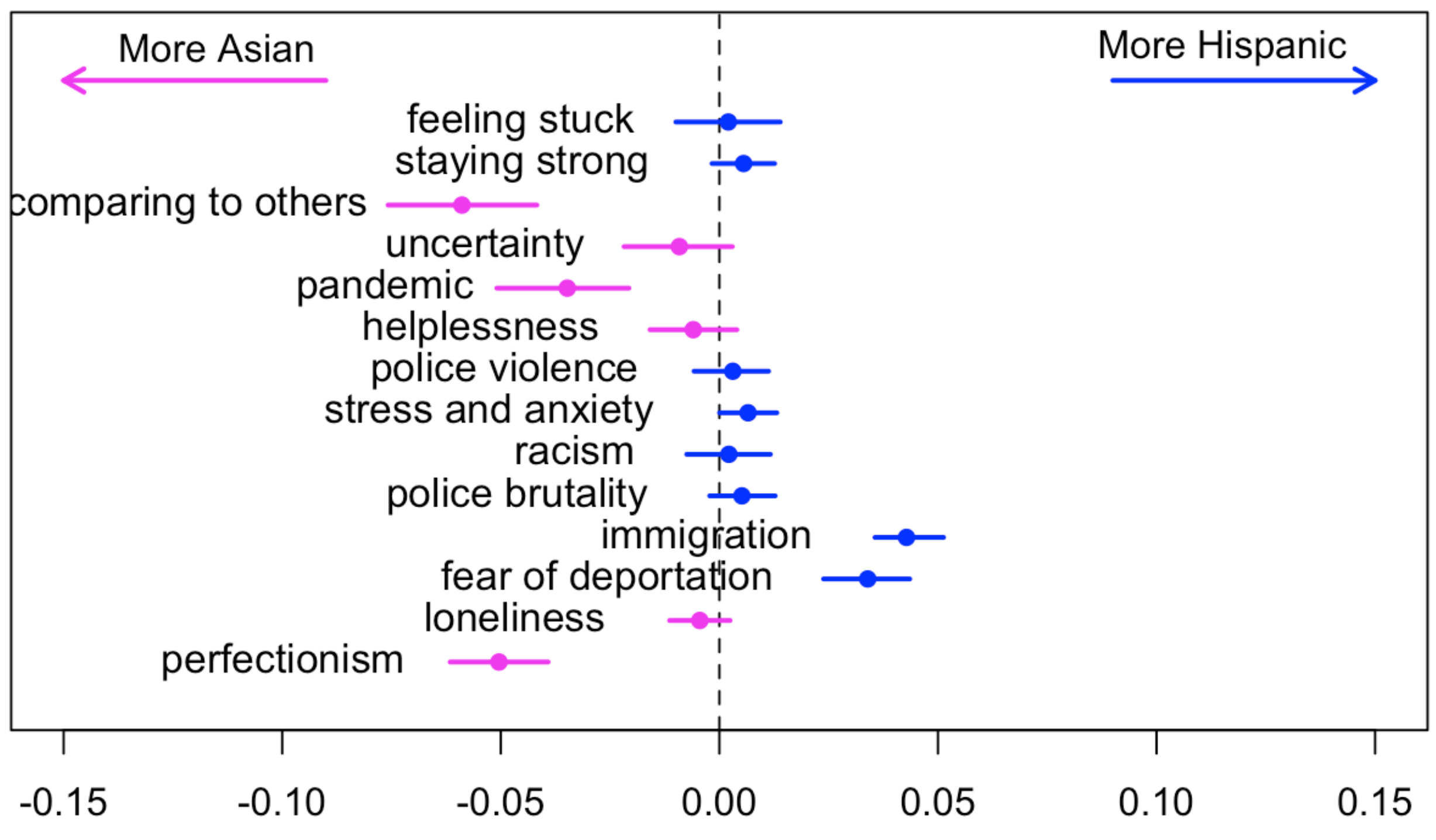} \\
(a) Asian vs. African American & (b) Asian vs. Hispanic \\[10pt]
 \includegraphics[width=70mm]{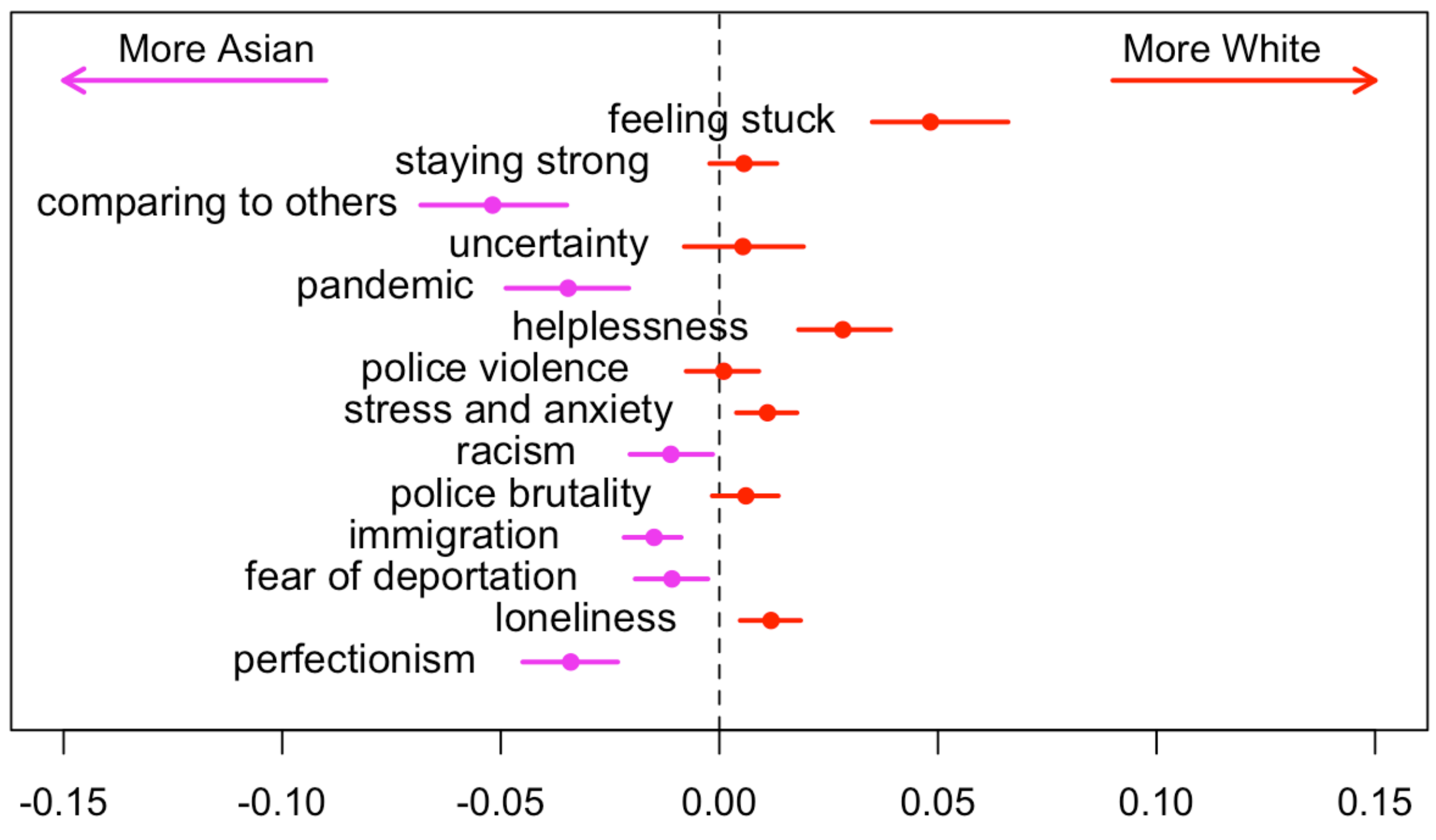} &   \includegraphics[width=70mm]{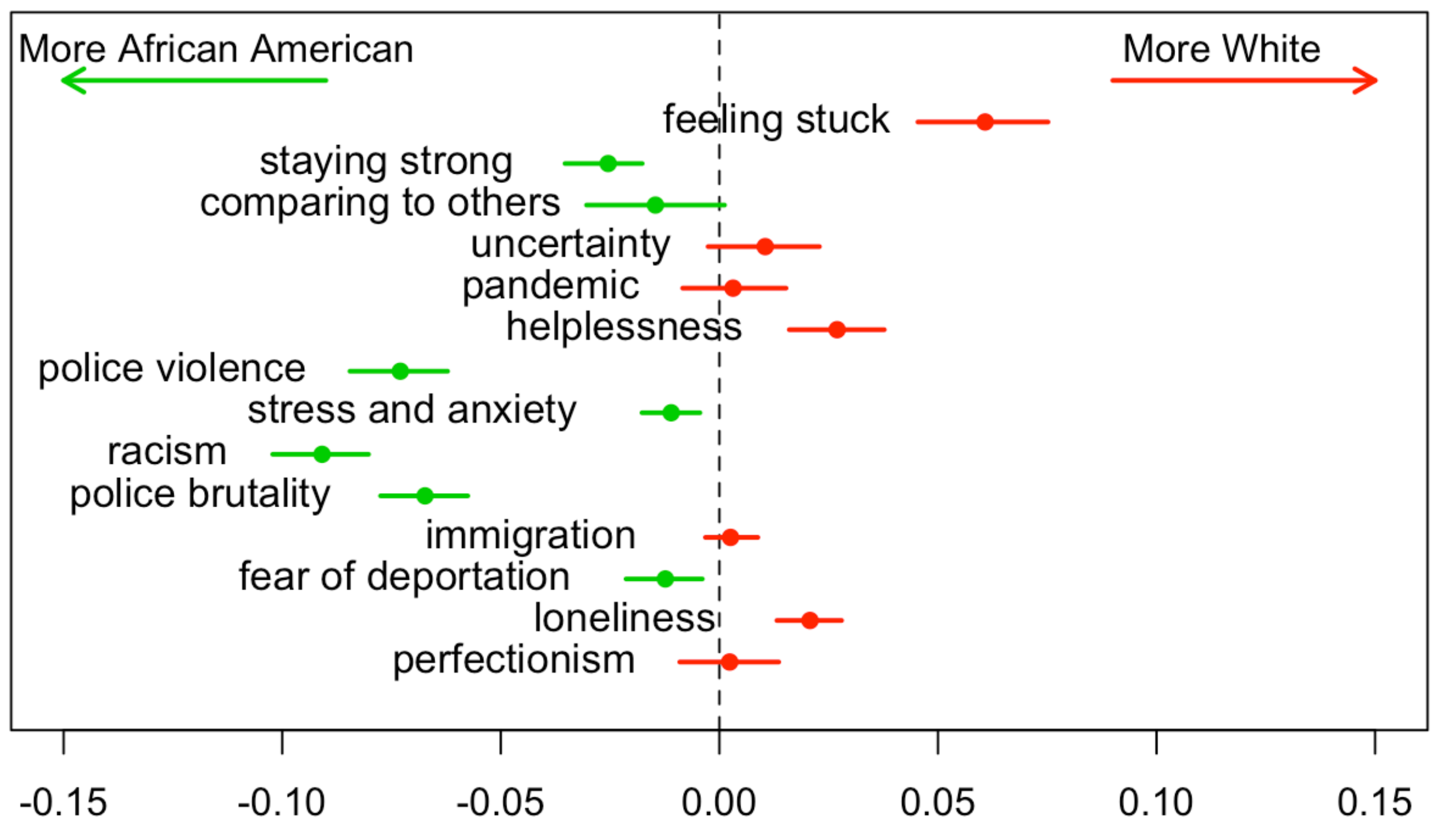} \\
(c) Asian vs. White & (d) African American vs. White \\[10pt]
\includegraphics[width=70mm]
{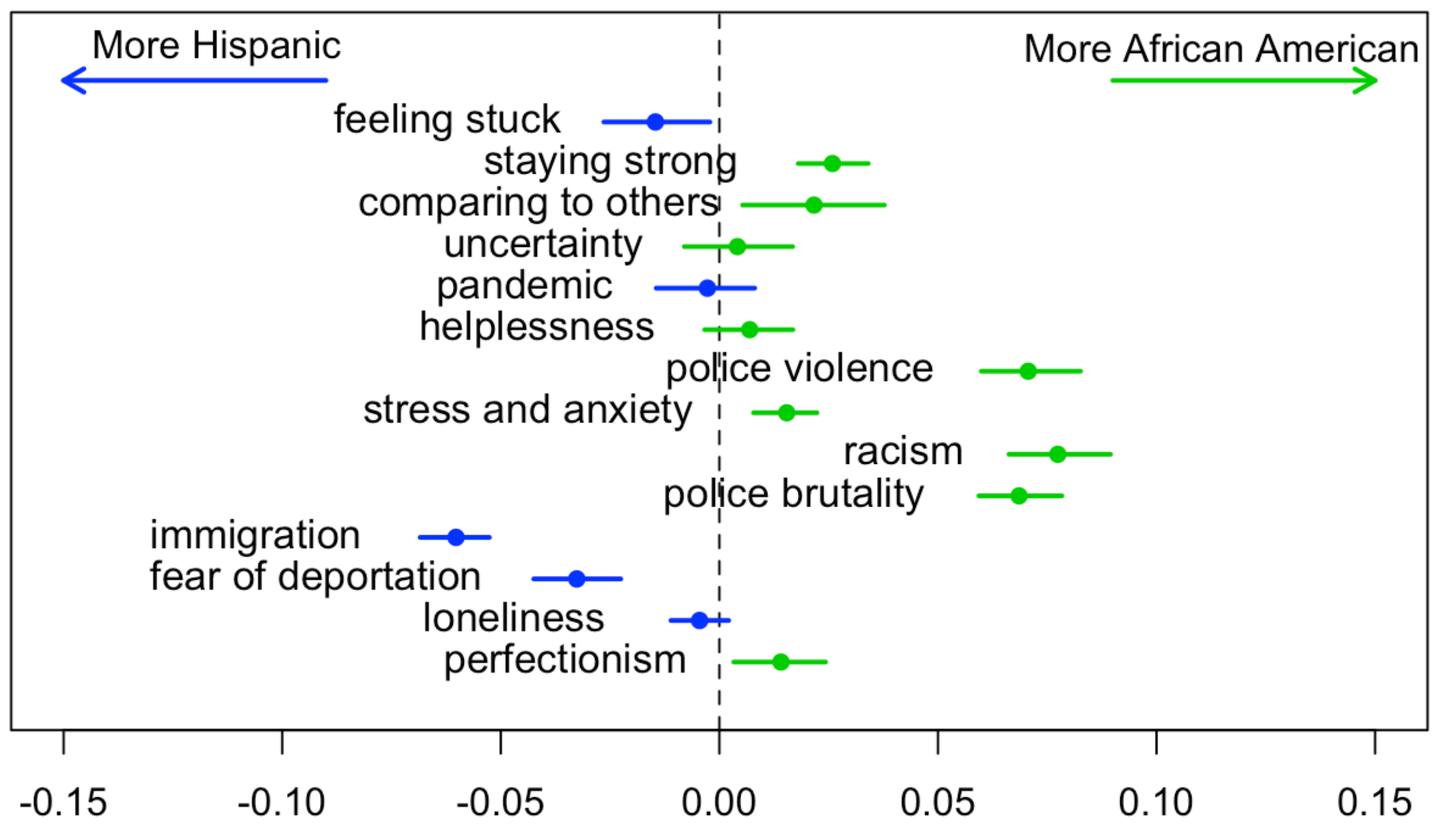} &   \includegraphics[width=70mm]{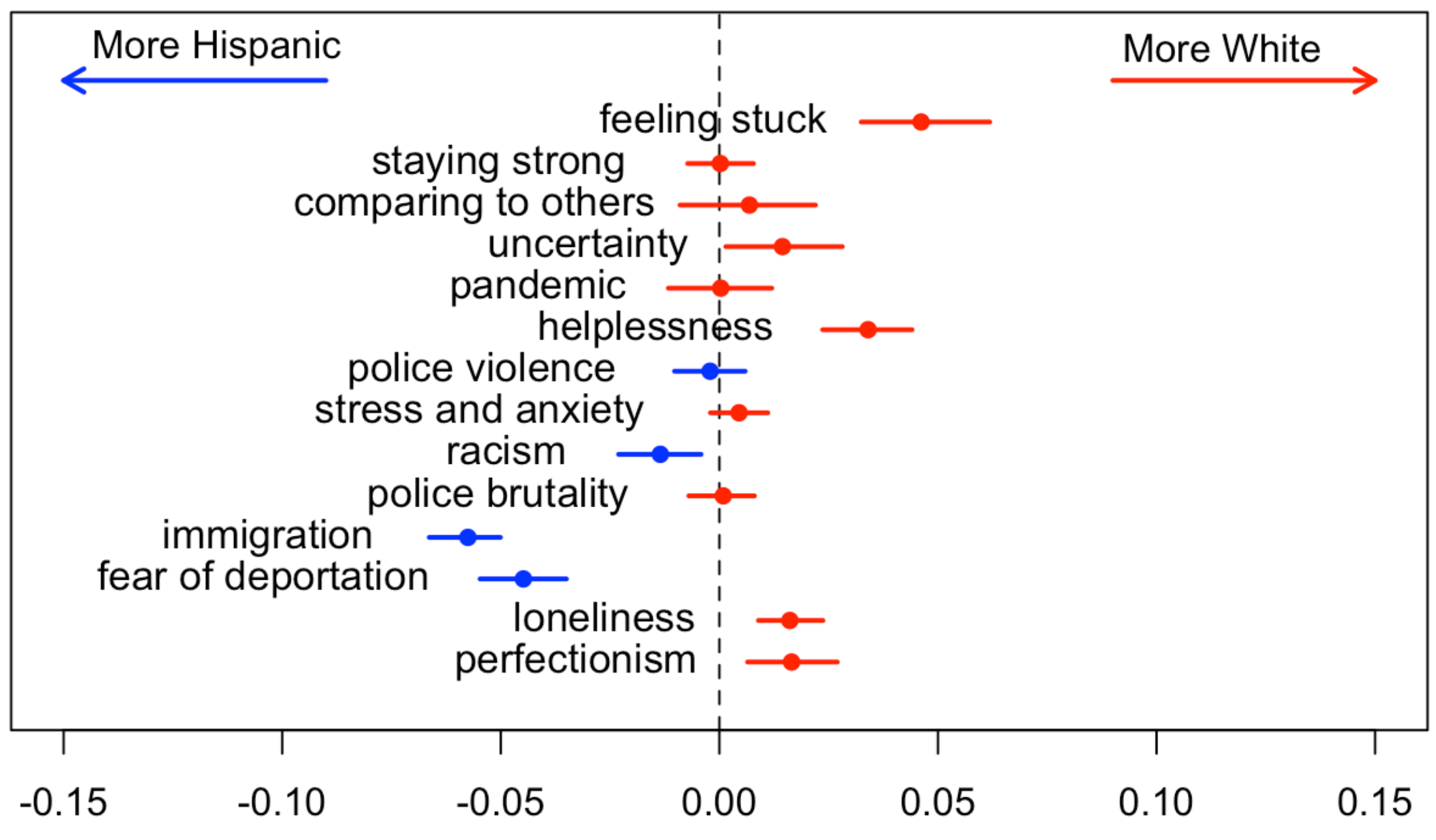} \\
(e) Hispanic vs. African American & (f) Hispanic vs. White \\[10pt]
\multicolumn{2}{c}{\includegraphics[width=70mm]{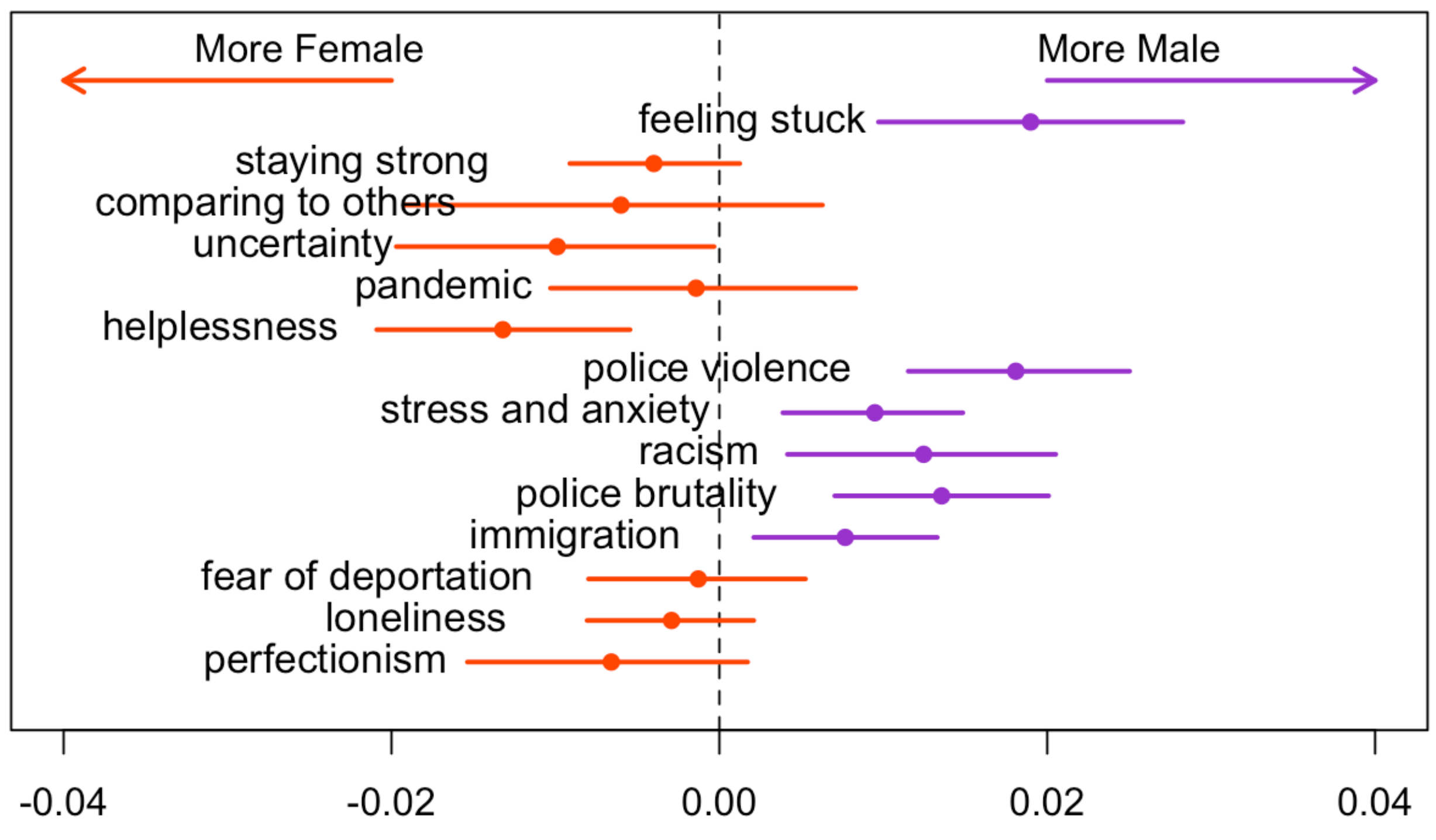} }\\
\multicolumn{2}{c}{(g) Women vs. Men}
\end{tabular}
\caption{Topic Modeling: topic proportion between different demographics, as detected in GPT-generated data and \textit{not} in real-life data. Colors represent different races and genders: Men -- purple, Women -- orange, Asian -- magenta, African American -- green, Hispanic -- blue, and White -- red. The bars represent confidence intervals. The closer to the graph extremities, the more prevalent the topics for the corresponding demographics.  \textit{Best viewed in color.}}
\clearpage


\label{fig:topic_diff_appendix}
\end{figure*}

The topic proportion between different demographics, and lexical analyses indicate demographic differences regarding stressors.  Refer to \Cref{fig:topic_diff}, \Cref{fig:topic_diff_appendix} and \Cref{tab:liwc-small} for the figures.
\vspace{-3mm}
\paragraph{Gender.} 
Between genders, women have more mentions of first-person pronouns (\texttt{pro1} and \texttt{ppron}) (\Cref{tab:liwc-small} (a)). 
Also, we find the following prevalent stressors: \texttt{health, news and social media, news and politics, family}, and \texttt{relationship}. See \Cref{fig:topic_diff} (g), \Cref{fig:topic_diff_appendix} (g).

In contrast, men tend to mention stressors regarding \texttt{finances and unemployment}, and \texttt{school} more than women. 
Furthermore, topics regarding \texttt{racism} and \texttt{police brutality} are much more prominent in men than women.

Both women and men mention stressors related to \texttt{work}, but for different reasons:  women about \texttt{work2/ work-pressure} and men about \texttt{work1/ work-fatigue}. The two types of work-related stressors are defined in Appendix \Cref{tab:topics}.

We acknowledge that we are excluding other gender identities by only comparing between two genders, \textit{women} and \textit{men}. 
We take this decision because the comparison data used in \citet{aguirre_gender_2021} is primarily from binary genders, \textit{women, men}, and very few from \textit{other}.  
\vspace{-3mm}

\paragraph{Race.}
We conduct a pairwise analysis for each race group.
\vspace{-3mm}

\paragraph{\textit{African American}.}%
The African American group tends to mention words related to 
\texttt{health, body, perception} and \texttt{family} (LIWC categories:  \texttt{bio, health, body, percept}, and \texttt{see}). 
Topics relating to \texttt{racism} and \texttt{police brutality} are also more likely compared to other groups.
See \Cref{tab:liwc-small} (b, e, f), \Cref{fig:topic_diff} (a, d, e), and \Cref{fig:topic_diff_appendix} (a, d, e) 
\vspace{-3mm}

\paragraph{\textit{Asian.}} 
For the Asian group, the topics \texttt{perfectionism} and \texttt{comparing to others} are significant stressors.
The Asian group also tend to be more concerned with \texttt{work, school} and \texttt{reward} (LIWC categories: \texttt{work, reward} and \texttt{achiev}). 
See \Cref{tab:liwc-small} (b, c, g), \Cref{fig:topic_diff} (a, b, c), and \Cref{fig:topic_diff_appendix} (a, b, c).
\vspace{-3mm}

\paragraph{\textit{Hispanic.}} 
The Hispanic group has more stressors related to \texttt{immigration}.
Other stressors include \texttt{family} and \texttt{social interactions}, while other prominent topics include \texttt{finances, news, work}. 
See \Cref{tab:liwc-small} (d, f, g), \Cref{fig:topic_diff} (b, e, f), and \Cref{fig:topic_diff_appendix} (b, e, f).
\vspace{-3mm}

\paragraph{\textit{White.}} 
In the White group, the most prominent stressors are: \texttt{general stress, news and social media, relationships} and \texttt{uncertainty}.
See \Cref{tab:liwc-small} (c, d, e), \Cref{fig:topic_diff} (c, d, f), and \Cref{fig:topic_diff_appendix} (c, d, f).

\begin{figure}
    \centering
    \includegraphics[width=70mm]{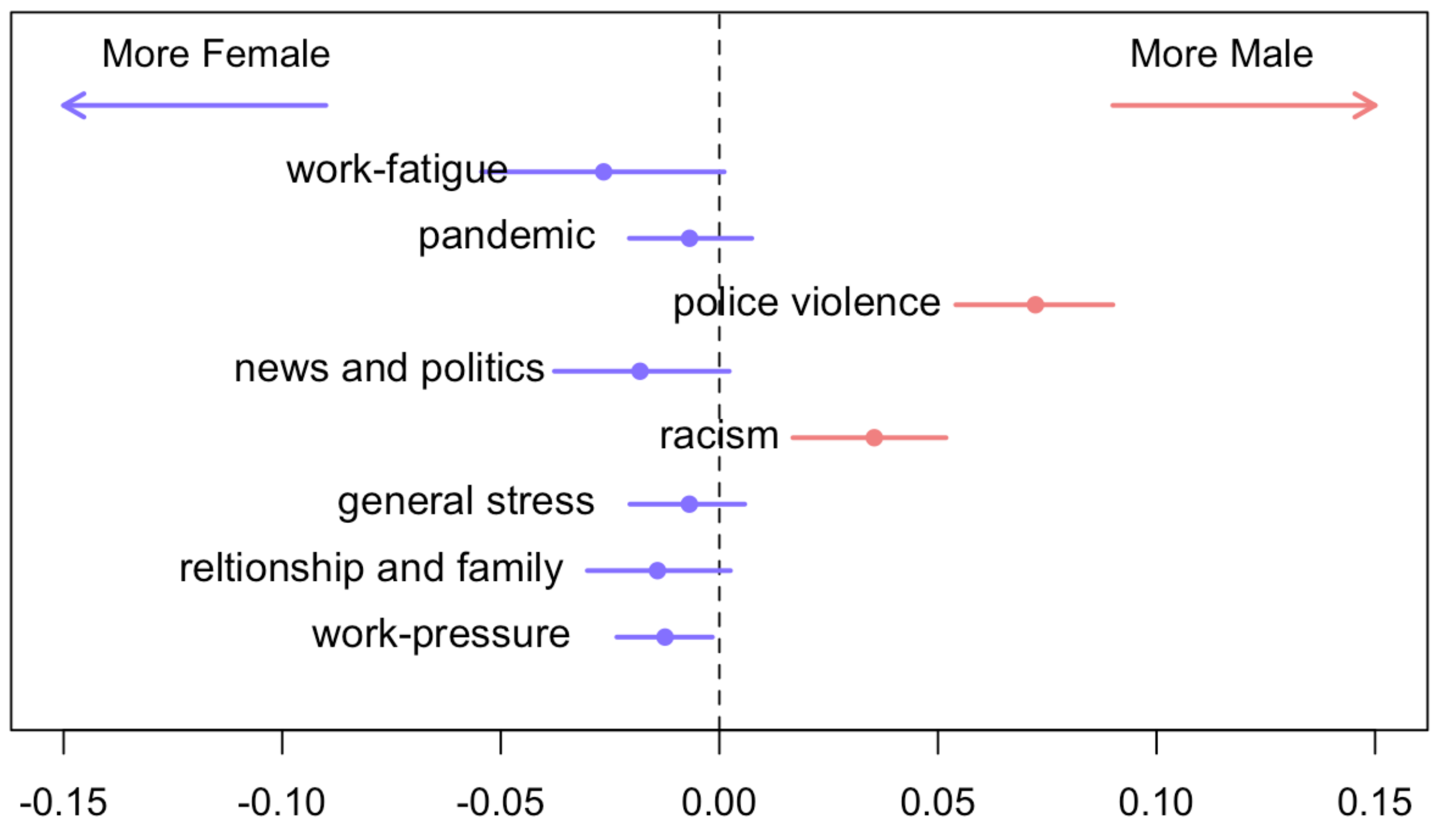}
    \caption{Topic Modeling: topic proportions between race and gender intersectionality -- African American women vs. African American men. The bars represent confidence intervals. The closer to the graph extremities, the more prevalent the topics are for the corresponding demographics
    }
    \label{fig:bw_comp}
\end{figure}

\vspace{-3mm}
\paragraph{Race and Gender Intersectionality.}

We also analyze the intersectionality of race and gender in \textsc{HeadRoom}, and provide an excerpt of the experiment to demonstrate how it could be used to study the data further.
Analyzing all possible demographic combinations would be too expansive, hence we only provide an excerpt to demonstrate its use case.
Focusing on only one demographic category, such as race or gender, can overlook the fine-grained inequalities in demographic groups.
\citet{field_survey_2021} give the example that only looking at African American Group emphasizes the more gender-privileged group (African American men), and similarly, only looking at gender may lead to over-representing the race-privileged group (White women).
We fit an STM model on the race-gender metadata to find stressor patterns comparing \textbf{African American women} to \textbf{African American men}.
In \Cref{fig:bw_comp}, we show that \texttt{police violence} and \texttt{racism} are more prevalent stressors for African American men.  African American women are more concerned about \texttt{work, pandemic, news and politics, general stress}, and \texttt{relationship and family}.

In future work, if we can access a demographically labeled depression dataset, we can compare the intersectional stressor patterns to real-life data, as the data from \citet{aguirre_using_2022} does not include analyses on intersectionality.




\paragraph{RQ2. How does synthetic data about depression stressors compare to human-generated data across demographics?}
We compare the analysis findings from our synthetic dataset with the findings from UMD-ODH ~\cite{aguirre_using_2022}: (1) between the \textit{keywords} extracted from the aggregated data, not split by demographics,\footnote{We compute over the aggregated data as we could not obtain keywords split by demographic from \citet{aguirre_using_2022}.} and (2) between the stressors obtained for each demographic group. 
When comparing the stressors across demographics, we also compare them with other findings related to stressor patterns~\citep{mcknight-eily_racial_2021, aguirre_using_2022, loveys_cross-cultural_2018}.
Our analysis results as depicted in \Cref{fig:topic_diff}, and \Cref{tab:liwc-small} show that the most prevalent depression stressors across demographics are comparable between the human-generated and the synthetic datasets.
At the same time, GPT-3 also identifies other stressors not present in the real-life data, as shown in \Cref{fig:topic_diff_appendix}.
\vspace{-3mm}

\paragraph{Topic Similarity.}
From \citet{aguirre_using_2022}, we obtain the top 30 most prevalent keywords for each topic from UMD-ODH. We then compare them with the keywords from our topics to measure how closely they match each other.

For each topic pair, we convert all keywords in each topic into word embeddings using GloVe ~\cite{pennington-etal-2014-glove}.\footnote{We use \texttt{glove.6B.300d} from \url{https://nlp.stanford.edu/projects/glove/}}
We then average the embeddings within each topic, and calculate the cosine similarity of the averaged embeddings.
Topics with one-to-many matches (e.g., \texttt{work1/ work-fatigue} and \texttt{work2/ work-pressure}) are consolidated into one topic.
\Cref{tab:keywords} shows the cosine similarity scores between each topic pair.
\vspace{-3mm}

\paragraph{Gender.}
Comparing gender-related stress patterns, the findings in \citet{aguirre_using_2022} show that stressors related to \texttt{finances}, \texttt{relationships}, and \texttt{health} are more prevalent for women than men. In contrast, stressors about \texttt{social interactions} (LIWC categories: \texttt{home, leisure, social, affiliation}, \texttt{we} and \texttt{family}) are more prevalent in men. 

Our findings largely support this pattern and show that the prevalence of \texttt{health} and \texttt{relationships} stressors are more dominant in women.

However, different from \citet{aguirre_using_2022}, we find that in \textsc{HeadRoom}, stressors related to \texttt{finance} are more dominant in men than in women (LIWC category: \texttt{money}).
See \Cref{tab:liwc-small} (a), \Cref{fig:topic_diff} (g), and \Cref{fig:topic_diff_appendix} (g).

\vspace{-3mm}

\paragraph{African American Group.} 
Real-life depression patterns indicate that African Americans are more likely to discuss \texttt{health} and use more \texttt{social} terms compared to other groups~\cite{loveys_cross-cultural_2018}. 
Our synthetic data also supports this: 
The stressor \texttt{health} is more prevalent for African American groups than for Asian, White, or Hispanic groups (LIWC categories: 
\texttt{health, body}, and \texttt{bio}).


In our synthetic data, we also find that stressors for \texttt{police brutality}, \texttt{police violence}, and \texttt{racism} are more prominent in African American groups despite these topics not being present in \citet{aguirre_using_2022}.
However, \citet{alang_police_2021} showed that hostile police encounters significantly affect African American and Hispanic individuals and are associated with depressed mood and anxiety.
See \Cref{tab:liwc-small} (b, e, f), \Cref{fig:topic_diff} (a, d, e), and \Cref{fig:topic_diff_appendix} (a, d, e).
\vspace{-3mm}

\paragraph{Asian Group.} 
In our synthetic data, the Asian Group demonstrates a more substantial prevalence of \texttt{school} stressors, matching prior findings from \citet{aguirre_gender_2021, loveys_cross-cultural_2018}.
Surprisingly, topics relating to the impacts of COVID-19 are more commonly associated with the Asian group, despite prior findings showing that Hispanic groups were severely affected by it due to lack of housing and basic needs~\citep{mcknight-eily_racial_2021}.
See \Cref{tab:liwc-small} (b, c, g), and \Cref{fig:topic_diff} (a, b, c).
\vspace{-3mm}

\paragraph{Hispanic Group.} 

\citet{aguirre_using_2022, mcknight-eily_racial_2021} showed that stressors \texttt{education}, \texttt{finance}, \texttt{government}, and \texttt{family} are prevalent in the Hispanic group.
These findings align with ours: 
\texttt{finance}, \texttt{school}, \texttt{family}, and \texttt{politics} are more prevalent for the Hispanic group than other groups 

However, \citet{loveys_cross-cultural_2018} found that the Hispanic group tends to make fewer mentions of social terms than African Americans, which contradicts our findings.
We find that LIWC categories \texttt{social} and \texttt{affiliation} are more common in Hispanic groups.
See \Cref{tab:liwc-small} (d, f, g), \Cref{fig:topic_diff} (b, e, f), and \Cref{fig:topic_diff_appendix} (b, e, f).
\vspace{-3mm}

\paragraph{White Group.} 
Similar to \citet{aguirre_using_2022} and \citet{mcknight-eily_racial_2021}, we find that for the White Group, the \texttt{finance} stressor is less prevalent than in other racial groups (\Cref{fig:topic_diff} (c, d, f)).

Different from previous findings~\citep{mcknight-eily_racial_2021}, when comparing White and African American groups, we find that \texttt{family} stressors are more prevalent in the African American group (see \Cref{fig:topic_diff} (d)).

\section{Conclusion}
In this paper, we developed a procedure to produce depression data using GPT-3, which could be applied to other LLMs to test their capability for creating synthetic mental health data.  We perform semantic and lexical analyses on this dataset to understand how GPT-3 represents depression stressors across demographics.  Our findings show the differences in the types of depression stressors GPT-3 attributes to different demographics, and that some prominent stressors across demographics are similar to those in real-life data from UMD-ODH.
Our synthetic data and code is for research purposes only and is made available at \url{https://github.com/MichiganNLP/depression_synthetic_data}.

\section{Acknowledgments}

We thank the anonymous reviewers for their constructive feedback. This project was partially funded by a National Science Foundation award (\#2306372). Any opinions, findings, and conclusions or recommendations expressed in this material are those of the authors and do not necessarily reflect the views of the National Science Foundation. 
We thank Philip Resnnik and Carlos Aguirre for providing us with their topic model generation code as well as the data.


\section*{Ethical Statement and Limitations}
\subsection{Ethical Statement}
We clarify that the intent of our research nor our dataset is not a proxy for creating mental health datasets.
We see our paper as a way to discover the biases that LLMs have for different demographics and compare them with available human data. We do not believe this data should be used for supplementing current human data because it can enforce biases. Instead, we propose to use our data for research, to investigate the biases of current LLMs in mental health and how they compare to human data.
Our dataset is created using only GPT-3, and according to the IRB of our institution, does not classify as a human subjects research.
It is also difficult to explain why GPT-3, or LLMs in general, speculate these stressors, and lack of explainability of the outputs should be considered when following our methods.  However, despite the lack of explainability, the recent evolution in the quality of LLM output is driving researchers to consider its application in synthetic data generation such as hate-speech data\cite{moller_is_2023}. Creating procedures to analyze the algorithmic fidelity of these datasets encourage researchers to use LLMs for synthetic data generation with caution by developing a series of methods to understand its underlying behaviors and potential risks.
\subsection*{Limitations}
\paragraph{Gender Representation.}
We are aware that the gender and race categories we explored are exclusionary and do not capture the full spectrum of gender identity, sexuality, race, and ethnicity; our choice in race and gender groups were predominantly decided by the availability of existing, human-generated datasets.
\vspace{-3mm}

\paragraph{Location Representation.}
The authors who collected the UMD-ODH dataset did not mention the location of the patients, so we also do not mention it in the GPT-3 prompts~\cite{kelly_can_2021, kelly_blinded_2020}. The data it generates is probably not comprehensive of the whole world, and the findings do not represent all cultures.
\vspace{-3mm}

\paragraph{Sensitive Information.} 
Across all groups, we note that mentions of \textit{suicide} or \textit{self-harm} are not included in our synthetic data. At the same time, they tend to be mentioned in real-life depression texts~\citep{aguirre_using_2022}.
This difference may be a result of model restrictions.
\vspace{-3mm}

\paragraph{Dataset Size.}
The size of the dataset is based on the UMD-ODH dataset used by \cite{aguirre_using_2022} which consisted of 2607 samples; we also keep our synthetic dataset size small while balancing for demographic groups to conduct a fair comparison to their results.
\vspace{-3mm}
\paragraph{Using Real-life Depression Data.}
Due to the difficulty of obtaining demographically-labeled depression datasets, we could not conduct fine-grained analyses between our data and human-generated depression data.
While we conduct some quantitative analyses based on the topic keywords provided by the authors of \citet{aguirre_using_2022}, having access to a human-generated dataset would have allowed us to obtain more detailed observations.  
We also produced relatively short samples, and we do not know whether these stressor patterns hold for longer text sequences.

\vspace{-3mm}
\paragraph{Model Variability.}
At the moment, OpenAI does not mention updating \texttt{text-davinci-003}; however, we do not know if this will remain true.
Possible changes to the model may alter our findings.
Additionally, the model is only trained with data up to June 2021, and cannot predict relevant stressors beyond that time frame. 
The prompts used here are flexible in that the time context can be replaced easily to target a specific time frame, and can be used with other LLMs that has similar capabilities.  One could use another LLM model using our prompt to explore its potential to be applied in depression analysis.




\bibliographystyle{lrec-coling2024-natbib}
\bibliography{references3}

\newpage

\clearpage
\appendix

\section{Appendix}
\label{sec:appendix}

\begin{table}[h] \small
\begin{tabular}{ c | c}
Overarching Topics & Fine-grained Topic \\
\midrule
\multirow{2}{*}{\textbf{Work}} & work-fatigue (work 1) \\ & work-pressure (work 2)\\
\midrule
\multirow{3}{*}{Racism/ police brutality} & Fear of police and violence \\
& Racism\\
& Police brutality\\ 
\midrule
\multirow{8}{*}{General stress} & Feeling stuck \\ 
& Staying strong \\
& Uncertainty \\
& Comparing to others \\
& Helplessness \\
& Stress and anxiety \\
& Loneliness\\
& Perfectionism \\
\midrule 
\multirow{2}{*}{Immigration status} & Fear of deportation \\
& Life as an immigrant\\
\midrule 
\multirow{2}{*}{\textbf{News}} & News and social media\\
& Politics and economy\\
\midrule 
\multirow{1}{*}{\textbf{Finances}} &  Finances and unemployment \\
\midrule 
\multirow{1}{*}{Pandemic} & Pandemic\\
\midrule 
\multirow{1}{*}{\textbf{Family}} & Family \\
\midrule 
\multirow{1}{*}{\textbf{Relationships}} & Relationships \\
\midrule 
\multirow{1}{*}{\textbf{Health}} & Health\\
\midrule 
\multirow{1}{*}{\textbf{School}} & School\\
\midrule 
\end{tabular}
    \caption{All topics from our synthetic data. The overarching topics that also match the topics in the UMD-ODH data are highlighted in \textbf{bold}.}
    \label{tab:topics}
\end{table}

\begin{table}[!ht]
    \centering
    \begin{tabular}{@{\extracolsep{4pt}}l l l l@{}}
    \toprule
    \multicolumn{4}{ c }{Gender} \\
    \hline
     \multicolumn{2}{c}{Women(+)} & \multicolumn{2}{c}{Men(-)}\\
    \hline
        category & ratio & category & ratio\\ 
    \cmidrule(lr){1-2}\cmidrule(lr){3-4}
      
        female & 3.95 & \textbf{male} & -4.06 \\ \hline
        adverb & 3.16 & \textbf{see} & -2.45 \\ \hline
        i & 2.83 & we & -1.88 \\ \hline
        \textbf{pro1} & 2.56 & verb & -1.76 \\ \hline
        feel & 1.81 & ipron & -1.70 \\ \hline
        \textbf{anx} & 1.52 & auxverb & -1.35\\ \hline
        ppron & 1.32 & tentat & -1.21 \\ \hline
        affect & 1.27 & body & -1.21\\ \hline
        posemo & 1.26 & article & -1.05 \\ \hline
        home & 1.10 & money & -1.02 \\ \hline
        insight & 1.06 & interrog & -1.01 \\ \hline
        leisure & 1.04 & health & -0.95 \\ \hline
        \textbf{sad} & 1.02 & compare & -0.89 \\ \hline
        friend & 1.00 & focuspast & -0.86 \\ \hline
        conj & 0.99 & discrep & -0.84\\ \hline
    \end{tabular}
    \caption{Lexical analysis of our data between Women and Men. LIWC categories that also matches the topics in the UMD-ODH data are highlighted in \textbf{bold}}
    \label{tab:liwc_wm}
\end{table}

\begin{table}[!ht]
    \centering
    \begin{tabular}{l l l l }
    \toprule
    \multicolumn{4}{ c }{Ethnicity} \\
    \hline
     \multicolumn{2}{c}{Asian(+)} & \multicolumn{2}{c}{White}\\
       \hline
        category & ratio & category & ratio\\
     \cmidrule(lr){1-2}\cmidrule(lr){3-4}
        leisure & 2.73 & \textbf{anx} & -2.60\\ \hline
        family & 2.62& adverb & -2.57 \\ \hline
        certain & 2.60 & see & -2.54 \\ \hline
        pro1 & 2.57 & motion & -1.84 \\ \hline
        home & 2.50 & \textbf{negemo} & -1.55 \\ \hline
        work & 2.24 & focusfuture & -1.49 \\ \hline
        i & 2.20 & interrog & -1.44 \\ \hline
        reward & 2.14 & tentat & -1.37\\ \hline
        achiev & 1.62 & ingest & -1.32 \\ \hline
        drives & 1.59 & insight & -1.16 \\ \hline
        posemo & 1.47 & space & -1.02\\ \hline
        negate & 1.34 & relativ & -0.99 \\ \hline
        auxverb & 1.14& anger & -0.98 \\ \hline
        focuspast & 1.09 & percept & -0.92 \\ \hline
        nonflu & 1.08 & adj & -0.86\\ \hline
    \end{tabular}
     \caption{Lexical analysis of our data between Asian and White group. LIWC categories that also matches the topics in the UMD-ODH data are highlighted in \textbf{bold}}
    \label{tab:liwc_aw}
\end{table}

\begin{table}[!ht]
    \centering
    \begin{tabular}{ l l l l }
    \hline
     \multicolumn{2}{c}{Hispanic(+)} & \multicolumn{2}{c}{White(-)}\\
       \hline
        category & ratio & category & ratio\\
     \cmidrule(lr){1-2}\cmidrule(lr){3-4}
        home & 7.48 & insight & -3.36 \\ \hline
        leisure & 7.37 & percept & -3.17\\ \hline
        family & 7.18 & cogproc & -2.95 \\ \hline
        affiliation & 5.30 & see & -2.62\\ \hline
        we & 4.36 & feel & -2.62\\ \hline
        drives & 2.63& tentat & -2.39\\ \hline
        social & 2.51 & compare & -1.83 \\ \hline
        focuspast & 2.28 & differ & -1.74 \\ \hline
        money & 2.25 & ipron & -1.48 \\ \hline
        anx & 2.01 & \textbf{health} & -1.28 \\ \hline
        achiev & 1.75 & \textbf{bio} & -1.23 \\ \hline
        auxverb & 1.53 & space & -1.22 \\ \hline
        cause & 1.44 & power & -1.12 \\ \hline
        \textbf{number} & 1.30 & prep & -1.10 \\ \hline
        pro1 & 1.07 & negate & -1.05 \\ \hline
    \end{tabular}
    \caption{Lexical analysis of our data between Hispanic and White group. LIWC categories that also matches the topics in the UMD-ODH data are highlighted in \textbf{bold}}
    \label{tab:liwc_hw}
\end{table}

\begin{table}[!ht]
    \centering
    \begin{tabular}{ l l l l }
    \hline
     \multicolumn{2}{c}{African American(+)} & \multicolumn{2}{c}{White(-)}\\
       \hline
        category & ratio & category & ratio\\
     \cmidrule(lr){1-2}\cmidrule(lr){3-4}
        see & 5.88 & insight & -2.28\\ \hline
        bio & 4.01 & adverb & -2.25 \\ \hline
        \textbf{percept} & 3.78 & tentat & -2.07 \\ \hline
        certain & 3.74 & nonflu & -2.04 \\ \hline
        we & 3.68 & work & -1.81 \\ \hline
        \textbf{number} & 3.44 & informal & -1.81 \\ \hline
        health & 3.40 & you & -1.80 \\ \hline
        body & 2.78 & space & -1.78 \\ \hline
        time & 2.18 & ppron & -1.67 \\ \hline
        adj & 1.86 & power & -1.55 \\ \hline
        feel & 1.77 & differ & -1.45 \\ \hline
        compare & 1.75 & cogproc & -1.41 \\ \hline
        \textbf{prep} & 1.66 & shehe & -1.26 \\ \hline
        affiliation & 1.66 & discrep & -1.25 \\ \hline
        money & 1.62 & focusfuture & -1.18 \\ \hline
    \end{tabular}
    \caption{Lexical analysis of our data between African American and White group. LIWC categories that also matches the topics in the UMD-ODH data are highlighted in \textbf{bold}}
    \label{tab:liwc_bw}
\end{table}

\begin{table}[!ht]
    \centering
    \begin{tabular}{ l l l l }
    \hline
     \multicolumn{2}{c}{Hispanic(+)} & \multicolumn{2}{c}{African American(-)}\\
       \hline
        category & ratio & category & ratio\\
     \cmidrule(lr){1-2}\cmidrule(lr){3-4}
        home & 7.33 & see & -8.49 \\ \hline
        family & 6.79 & percept & -6.96 \\ \hline
        \textbf{leisure} & 6.72 & bio & -5.24 \\ \hline
        affiliation & 3.67 & health & -4.68 \\ \hline
        social & 3.51 & feel & -4.40 \\ \hline
        \textbf{anx} & 2.88 & compare & -3.59 \\ \hline
        focuspast & 2.81 & certain & -3.36 \\ \hline
        ppron & 2.47 & \textbf{prep} & -2.77 \\ \hline
        work & 2.05 & body & -2.74 \\ \hline
        you & 1.85 & adj & -2.56 \\ \hline
        drives & 1.76 & \textbf{number} & -2.14 \\ \hline
        focusfuture & 1.74 & ipron & -2.01 \\ \hline
        achiev & 1.66 & cogproc & -1.56 \\ \hline
        nonflu & 1.58 & time & -1.50 \\ \hline
        \textbf{informal} & 1.47 & quant & -1.17 \\ \hline
    \end{tabular}
    \caption{Lexical analysis of our data between Hispanic and African American group. LIWC categories that also matches the topics in the UMD-ODH data are highlighted in \textbf{bold}}
    \label{tab:liwc_hb}
\end{table}

\begin{table}[!ht]
    \centering
    \begin{tabular}{ l l l l }
    \hline
     \multicolumn{2}{c}{Hispanic(+)} & \multicolumn{2}{c}{Asian(-)}\\
       \hline
        category & ratio & category & ratio\\
     \cmidrule(lr){1-2}\cmidrule(lr){3-4}
        affiliation & 5.13 & cogproc & -3.00 \\ \hline
        we & 5.06 & feel & -2.99 \\ \hline
        home & 5.04 & i & -2.75 \\ \hline
        leisure & 4.70 & reward & -2.60 \\ \hline
        \textbf{family} & 4.61 & negate & -2.39 \\ \hline
        \textbf{anx} & 4.61 & percept & -2.26 \\ \hline
        social & 2.73 & certain & -2.23 \\ \hline
        \textbf{negemo} & 2.44 & insight & -2.21 \\ \hline
        focusfuture & 2.05 & work & -2.00\\ \hline
        adverb & 1.69 & posemo & -1.94 \\ \hline
        \textbf{money} & 1.57 & \textbf{compare} & -1.91 \\ \hline
        motion & 1.22 & nonflu & -1.53 \\ \hline
        interrog & 1.19 & pro1 & -1.50 \\ \hline
        focuspast & 1.19 & power & -1.34 \\ \hline
        verb & 1.14 & differ & -1.09 \\ \hline
    \end{tabular}
        \caption{Lexical analysis of our data between Hispanic and Asian group. LIWC categories that also matches the topics in the UMD-ODH data are highlighted in \textbf{bold}}

    \label{tab:liwc_ha}
\end{table}

\begin{table}
    \centering
    \begin{tabular}{ l l l l }
    \hline
     \multicolumn{2}{c}{Asian (+)} & \multicolumn{2}{c}{African American (-)}\\
       \hline
        category & ratio & category & ratio\\
       \cmidrule(lr){1-2}\cmidrule(lr){3-4}
        \textbf{work} & 4.06 & see & -8.41 \\ \hline
        i & 3.17 & bio & -4.79 \\ \hline
        nonflu & 3.11 & percept & -4.71 \\ \hline
        ppron & 2.64 & we & -4.38 \\ \hline
        home & 2.31 & health & -3.89\\ \hline
        family & 2.19 & body & -3.16 \\ \hline
        \textbf{leisure} & 2.03 & \textbf{number} & -2.90 \\ \hline
        power & 1.76 & adj & -2.73 \\ \hline
        negate & 1.74 & \textbf{prep} & -2.14 \\ \hline
        \textbf{informal} & 1.66 & risk & -1.93 \\ \hline
        focuspast & 1.62 & article & -1.91 \\ \hline
        reward & 1.57 & anx & -1.76 \\ \hline
        \textbf{achiev} & 1.53 & compare & -1.68 \\ \hline
        pronoun & 1.48 & ingest & -1.52 \\ \hline
        pro1 & 1.46 & affiliation & -1.48 \\ \hline
    \end{tabular}
    \caption{Lexical Analysis on our synthetic data: Log-odds-ratio of LIWC categories between Asians and African Americans. LIWC categories that also matches the topics in the UMD-ODH data are highlighted in \textbf{bold}}
    \label{tab:liwc_ab}
\end{table}

\end{document}